\documentclass[table, twocolumn,a4paper,twoside]{article}

\usepackage{authblk}  %To add authors with multiple affiliations

\usepackage{amsmath,amsfonts}
\usepackage[natbibapa]{apacite}%
\usepackage{apacdoc}

\bibpunct{(}{)}{; }{}{}{}%

\usepackage{hyperref}
\usepackage[]{graphicx}
\usepackage{longtable}
\usepackage{lscape}
\usepackage{multirow}
\usepackage{color}
\usepackage{array}
\usepackage{calc}
\usepackage{natbib}
\usepackage{enumitem}

\usepackage[english]{babel}

\usepackage{balance}
\usepackage{xcolor}
\makeatletter
\def\NAT@aysep{,}
\makeatother

\usepackage{titlesec}
\titleformat*{\section}{\centering \fontsize{10}{10} \bf}
\titleformat*{\subsection}{\fontsize{10}{10} \bf}

\begin{document}

\title{\textbf{Transfer Learning-based Road Damage Detection for Multiple Countries}}

\date{}

% \author{Deeksha Arya \footnote{deeksha@ct.iitr.ac.in}, Hiroya Maeda, Sanjay Kumar Ghosh, Durga Toshniwal, Alexander Mraz, \\Takehiro Kashiyama, Yoshihide Sekimoto\footnote{sekimoto@iis.u-tokyo.ac.jp}\\
% University of Tokyo, 4-6-1 Komaba, Tokyo, Japan}

\author[1,2]{\small Deeksha Arya\footnote{deeksha@ct.iitr.ac.in}}
\author[2]{Hiroya Maeda}
\author[1,3]{Sanjay Kumar Ghosh}
\author[1,4]{Durga Toshniwal}
\author[2,5]{Alexander Mraz}
\author[2]{Takehiro Kashiyama}
\author[2]{Yoshihide Sekimoto\footnote{sekimoto@iis.u-tokyo.ac.jp}}

\affil[1]{\footnotesize Centre for Transportation Systems (CTRANS), Indian Institute of Technology Roorkee, 247667, India}
\affil[2]{\footnotesize Institute of Industrial Science, University of Tokyo, 4-6-1 Komaba, Tokyo, Japan}
\affil[3]{\footnotesize Department of Civil Engineering, Indian Institute of Technology Roorkee, 247667, India}
\affil[4]{\footnotesize Department of Computer Science and Engineering, Indian Institute of Technology Roorkee, 247667, India}
\affil[5]{\footnotesize Amazon EU, 22 Rue Edward Steichen, 2540, Luxembourg}

% \author[1,2]{Deeksha Arya}

% \author[2]{Hiroya Maeda}

% \author[1,3]{Sanjay Kumar Ghosh}

% \author[1,4]{Durga Toshniwal}

% \author[2,5]{Alexander Mraz}

% \author[2]{Takehiro Kashiyama}

% \author[2]{Yoshihide Sekimoto}

% \authormark{Arya \textsc{et al}}

% \address[1]{\orgdiv{Centre for Transportation Systems (CTRANS)}, \orgname{Indian Institute of Technology Roorkee}, \orgaddress{\state{Uttarakhand - 247667}, \country{India}}}

% \address[2]{\orgdiv{Institute of Industrial Science}, \orgname{University of Tokyo}, \orgaddress{4-6-1 Komaba, \state{Tokyo}, \country{Japan}}}

% \address[3]{\orgdiv{Department of Civil Engineering}, \orgname{Indian Institute of Technology Roorkee}, \orgaddress{\state{Uttarakhand - 247667}, \country{India}}}

% \address[4]{\orgdiv{Department of Computer Science and Engineering}, \orgname{Indian Institute of Technology Roorkee}, \orgaddress{\state{Uttarakhand - 247667}, \country{India}}}

% \address[5]{\orgname{Amazon EU}, \orgaddress{\state{22 Rue Edward Steichen, 2540 }, \country{Luxembourg}}}

% \corres{Yoshihide Sekimoto\\ \email{sekimoto@iis.u-tokyo.ac.jp}}

\maketitle

\begin{abstract}
Many municipalities and road authorities seek to implement automated evaluation of road damage. However, they often lack technology, know-how, and funds to afford state-of-the-art equipment for data collection and analysis of road damages. Although some countries, like Japan, have developed less expensive and readily available Smartphone-based methods for automatic road condition monitoring, other countries still struggle to find efficient solutions.
This work makes the following contributions in this context. Firstly, it assesses usability of the Japanese model for other countries. Secondly, it proposes a large-scale heterogeneous road damage dataset comprising 26620 images collected from multiple countries using smartphones. Thirdly, we propose generalized models capable of detecting and classifying road damages in more than one country. Lastly, we provide recommendations for readers, local agencies, and municipalities of other countries when one other country publishes its data and model for automatic road damage detection and classification. Our dataset is available at (https://github.com/sekilab/RoadDamageDetector/).
\end{abstract}

% \begin{keywords}
% Automatic Road Condition Monitoring, Convolutional Neural Network, Deep Learning, Smartphone-based Road Damage Detection
% \end{keywords}

\footnotetext{\textbf{Abbreviations:} API, Application Programming Interface, CNN, Convolutional Neural Network, RDD, Road Damage Dataset, SSD, Single Shot Detector}

\section{Introduction}\label{Introduction}

Road infrastructure is a crucial public asset as it contributes to economic development and growth while bringing critical social benefits. It connects communities and businesses and provides access to education, employment, social, and health services. However, road surface wears and deteriorates over time from factors related to location, age, traffic volume, weather, engineering solutions, and materials being used to build it and, therefore, the knowledge of its deterioration extent are critical to efficient and cost-effective maintenance with the goal to preserve its good and safe condition. Many studies and surveys were made on the topic of roadway deficiencies and their impact on safety and economy (\cite{miller2009crash} and \cite{radopoulou2015detection}). 
Pavement distress, one of the pavement condition characteristics, is typically evaluated using one of three approaches: manual, semi-automated, or fully automated. The traditional methods for obtaining pavement condition data include manual and semi-automated surveys. In manual surveys, raters perform a visual inspection of the pavement surface through either walking on or along the pavement surface or by conducting a windshield survey from a slow-moving vehicle. Visual inspection of the road surface suffers from a subjective judgment of inspectors. It requires a significant human intervention that is proven to be time-consuming, given the extensive length of road networks. Moreover, inspectors must often be physically present in the travel lane, exposing themselves to potentially hazardous conditions. 
In semi-automated pavement condition evaluations, the road images are collected automatically from a fast-moving vehicle, but distress identification is postponed to an off-line process running in workstations at the office. This approach improves safety but still uses manual distress identification, which is very time-consuming. Currently, most state highway agencies use the semi-automated method (\cite{zalama2014road, pierce2013practical, mcghee2004automated}), which involves some degree of human intervention. Fully automated distress evaluations often employ vehicles equipped with sophisticated and expensive sensors. The processing of collected data is then conducted using image processing and pattern recognition software for distress identification and quantification. The data processing may be accomplished during data collection or in post-processing at the office. Raters’ role is to conduct quality assurance testing of the software functionality and perform quality control of the distress rating output by the software. Unlike the advanced data collection technology available to the transportation agencies, the pattern recognition software still needs further enhancements to detect and classify the various types of pavement surface distresses on different pavement surfaces to the accuracy levels acceptable to the agencies. Only a few highway agencies have implemented a fully automated crack detection system for network-level data collection. From experience, a systematic quality management process seems to be the central consideration for successful implementation (\cite{kargah2017evaluation}).
Specialized vehicles used for pavement inspection are usually equipped with multiple sensors such as laser scanners, road profilers, and cameras with the aim to capture road assets, including pavement images, and to acquire the longitudinal and transverse profiles of the road (\cite{Fugro2019}). However, such vehicles are expensive, and the cost to purchase such systems can reach half a million dollars, depending on the sensors included. Even though the operating price of road survey vehicles is cheaper than that of traditional surveying methods, it can reach between \$30 and \$50 per kilometer (\cite{radopoulou2015detection}). Meanwhile, mobile devices such as smartphones have evolved in recent years into devices containing high-resolution digital cameras, sensors, and powerful processors; therefore, examples of road inspection using smartphones are becoming more common. The advantage of employing smartphones is that they enable efficient and cost-effective inspection of the road surface of large road networks. For example, \cite{casas2016detection} proposed a method to visualize pothole detected by smartphone sensors on a map. \cite{mertz2014city} proposed a method to handle road images acquired by on-board smartphones installed on cars that operate daily, such as general passenger automobiles, buses, and garbage trucks, to detect road surface damage with an external laptop. \cite{maeda2018road} developed a smartphone application for the collection and real-time detection of roadway deficiencies used by Japanese local governments.
Even though Japanese Municipalities have started using the Smartphone-based model, there are many municipalities and local authorities in several other countries still lacking the technology, know-how, and funds to afford expensive state-of-the-art data collection equipment. On many occasions, they sustain their road networks through inefficient maintenance planning, and in some circumstances, they are unable to carry out adequate inspections. In the context of this challenging situation regarding the maintenance of such infrastructure, there is a need to assess the usability of the Japanese Road Damage Detection and Classification model for monitoring the road conditions in other countries. Further, it needs to be analyzed whether there would be any effect on the accuracy level of the Japanese model when used for other countries. If yes, what are the options available for other countries if the performance of the Japanese model is not found satisfactory when applied to their local roads? Do they need to train a new model for themselves from scratch, or can they utilize the Japanese model or data in some other way?
The above-mentioned questions need to be addressed based on some detailed research. Further, the recommendations are often sought for readers of other countries when one country, in this case, Japan, publishes its data or model for a specific application.
The presented work intends to answer the aforementioned research questions and provide the readers with apt recommendations regarding the automation of road damage detection and classification.
For the required comprehensive analysis, this study considers two countries, namely India and the Czech Republic (partially Slovakia), and conducts several experiments using different combinations of underlying data. 

First, localized road damage datasets were created by using 3590 road images from Czech and 9892 images from India using smartphones installed on the windshield of the vehicle. The newly collected datasets were labeled for cracks and potholes present in the images and were mixed with the Japanese dataset. 
Next, the study involves training and evaluating 16 deep neural network models considering 30 scenarios based on different combinations of the test and train datasets for detecting and classifying road damages. The evaluation results of these models are analysed to provide recommendations for the readers of other countries.

Overall, the contributions of this work can be listed as follows:

\begin{enumerate}[]
    \item	Performance analysis of the usability of the Japanese model for detecting and classifying the road damages in other countries.
    \item Proposing a large-scale heterogeneous dataset, comprising $26620$ annotated road images collected using a vehicle-mounted Smartphone from multiple countries.
    \item	Proposing generalized hybrid models capable of detecting and classifying road damages in more than one country. 
    \item	Providing recommendations for readers, local agencies/municipalities of other countries when one country publishes its data and model for automatic road damage detection and classification.
\end{enumerate}

The following section provides an overview of the existing literature.

\section{Related Work}\label{Related Work}

\subsection{Machine Learning and Convolutional Neural Networks}

\begin{figure*}[ht]
%\centerline{\includegraphics[width=389pt,height=16pc]{Fig_1_CNN_Arch.png}}
\centerline{\includegraphics[width=0.85\textwidth]{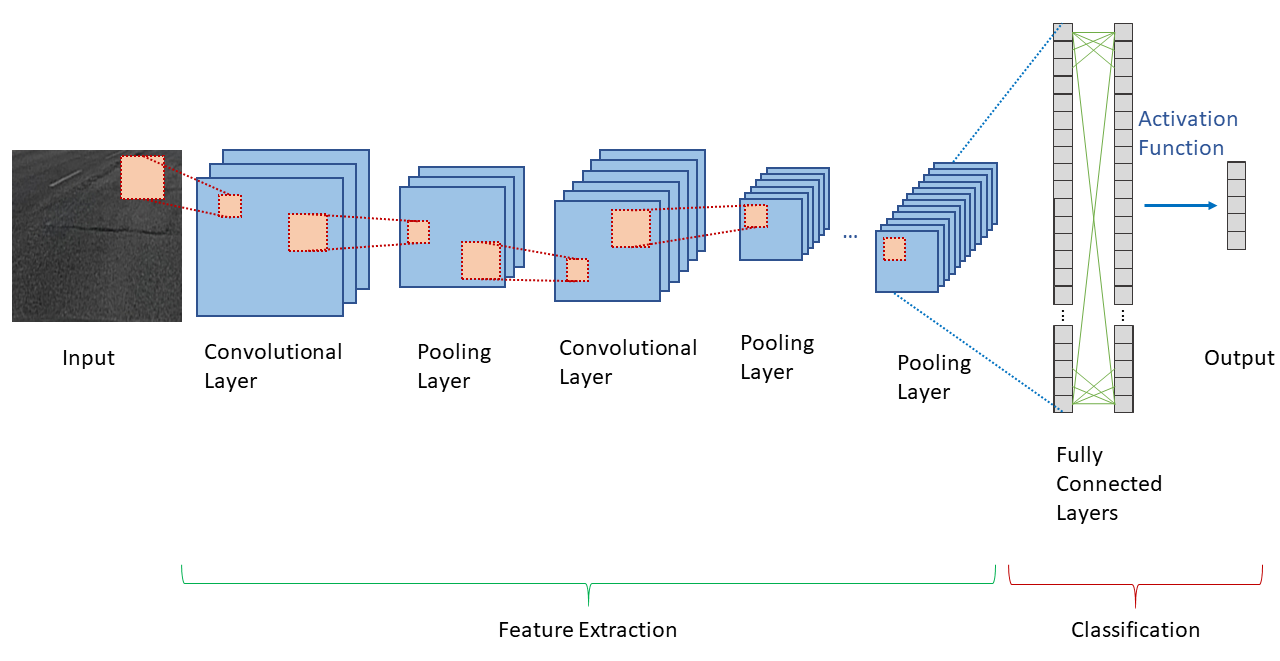}}
\caption{Typical CNN Architecture\label{fig1}}
\end{figure*}

The research area of neural networks is a successful field within computer science, which specializes in providing solutions in application domains that are difficult to model with conventional statistical approaches(\cite{adeli2001neural}). Such applications are usually characterized by noisy input data, largely unknown intrinsic structure, and changing conditions (\cite{hasenauer2001estimating}). The significant difference between neural networks and traditional methods of computer science is that the behavior of the neural networks is the result of a training process. In contrast, in traditional methods, the behavior of the system is predefined. Neural networks consist of artificial neurons that have learn-able weights and biases. Convolutional Neural Network (CNN) is one of the main categories of deep neural networks that are usually applied for image recognition and classification. These are the type of algorithms that can identify street signs, cars, faces, and many other types of objects. Each artificial neuron in CNN is activated after processing the input image using various convolutional operations, such as gradient, edge detection filters, blobs, in combination with learn-able weights and biases (\cite{peppa2018urban}). An example of a typical CNN architecture consists of series of convolution and pooling layers designed for feature extraction and fully connected layers and the activation function for classification of objects with probabilistic values ranging from 0 to 1, as shown in \ref{fig1}. The main advantage of CNN is that after the training stage, it automatically detects the critical features without any human supervision.

The core building block of CNN is a convolution layer that employs convolution operations that are applied to the input data using convolution filters, also called kernels, to produce an output called feature map. Convolution of an image with different filters is done for operations such as edge detection, blur, or sharpening. It uses small squares of input data to learn the image features and preserves the correlation between pixels (\cite{prabhu2018understanding}). As a result, the network determines which filters activate when the neural network detects some specific type of feature at some spatial position within the input image. Pooling layers are usually inserted between successive convolution layers, and their function is to progressively reduce the number of parameters in the network, which both shortens the training time and combats overfitting. It is a form of down-sampling. 
Contrary to convolution operation, pooling has no parameters, and it slides a window over its input and takes, for example, the maximum value in the windows (\cite{Dertat2017}). A fully connected layer performs high-level reasoning in the neural network. Neurons in the fully connected layer have a connection to all activations from neurons in the previous layer, like regular artificial neural networks. Activation function, also known as the transfer function, is used to map resulting values of the neural network, which may be negative or greater than one, into an interval between 0 and 1. For multi-class classification, a softmax function, which is a more generalized logistic activation function, is implemented (\cite{Sharma2017}). 

\begin{figure*}[ht]
\centerline{\includegraphics[width=389pt,height=10pc]{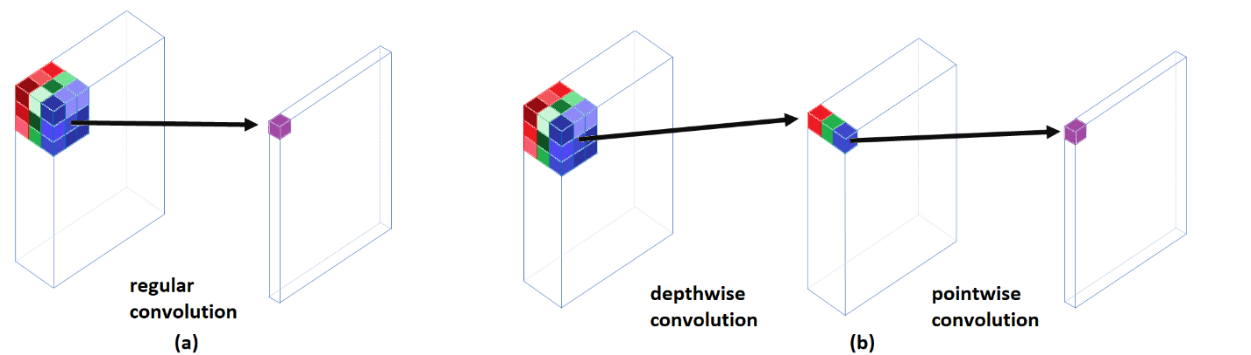}}

\caption{Comparison of (a) Regular and (b) Depth-wise Separable Convolutions applied on 3-channel (RGB) Image \label{fig2}}
\end{figure*}

Many CNN architectures, such as R-CNN (Region-based Convolutional Neural Networks;\cite{girshick2014rich}), Fast-CNN(\cite{girshick2015fast}), or Faster-CNN (\cite{ren2015faster}) have been developed to attain the best accuracy while improving the processing speed. However, the computational load was still too large for processing images on devices with limited computation, power, and space (\cite{he2016deep}). Therefore, the Single Shot Multi-Box Detector(SSD) framework was developed to improve further the computation speed of the object detection (\cite{liu2016ssd}). It uses a single feed-forward convolutional network to detect multiple objects within the image directly and combines predictions from numerous feature maps with different resolutions to handle objects of various sizes.
MobileNet is a small, low-latency, and low-power convolutional feature extractor that can be built to perform classification, detection, or segmentation similar to popular large-scale models, such as Inception SSD (\cite{szegedy2016rethinking}). It is based on depth-wise separable convolution, which factorizes a standard convolution into a depth-wise convolution, and $1\times1$ convolution called a point-wise convolution. Depth-wise convolution maps a single convolution on each input channel separately, and point-wise convolution is a convolution with a kernel size of $1\times1$ that combines the features created by the depth-wise convolution (\cite{Douillard2018}). In comparison to depth-wise separable convolution, a regular convolution does both filtering and combination steps in the single run. However, SSD MobileNet requires more computational work to accomplish the task, and it needs to learn more weights (Figure \ref{fig2}). More information on MobileNet is given in a paper published by \cite{howard2017mobilenets}. \cite{huang2017speed} compared many neural networks and object detection methods, and among them, the SSD object detection framework using a MobileNet base network was found to require relatively small CPU load and memory consumption while maintaining high accuracy. \cite{Budzar2018} compared the accuracy of the famous large-scale Inception model to MobileNet and concluded that the difference was less than 6\% (95.2\% vs. 89.5\%, respectively) while MobileNet was more than six-times smaller when compared to Inception model(13 MB vs. 87 MB). MobileNet has also been shown to achieve accuracy comparable to VGG-16 on the ImageNet dataset with only 1/30th of the computational cost and model size (\cite{maeda2018road}). MobileNet is designed to effectively maximize accuracy while being mindful of the restricted resources for on-device or embedded applications. Due to all these advantages, the study presented in this paper adopted SSD MobileNet.

\subsection{Road Infrastructure Inspection using Machine Learning}
Recent studies have adopted various deep learning neural network systems for automated road surface survey or damage detection(\cite{cha2018autonomous}, \cite{zhang2017automated}, \cite{lin2017structural}). For example, \cite{LealDeSilva2018concrete} developed a machine learning-based model to detect cracks on concrete surfaces. Supervised deep convolutional neural network was trained by \cite{zhang2016road} to classify pavement images taken by smartphones around the Temple University campus. However, proposed road damage detection methods focused only on the determination of the existence of damage. \cite{anand2018crack} presented the development of deep neural network architecture using GPU board and associated camera to detect road cracks and potholes for self-driving cars and autonomous robots to perform necessary evading maneuvers to ensure a smooth journey. Supervised, convolutional neural network (CNN) was trained by \cite{fan2018automatic} to recognize different pavement conditions, and results presented in the paper show that it can deal well with varying pavement textures. 
The detection method developed by \cite{maeda2018road} classified pavement deterioration of Japan road network into eight categories based on images captured by mobile devices with a resolution of 300 x 300 pixels. The corresponding dataset, named RDD-2018, was made publicly available in 2018, and a Smartphone-based application was also introduced for real-time road condition assessment. Since then, this application is being used by several municipalities in Japan for faster monitoring of road conditions. As a result, the underlying data, method, and models have gained wide attention from researchers all over the world. 
A technical challenge was organized in December 2018 as a part of the IEEE Big Data Conference held at Seattle, USA, which utilized this data for evaluating the performance of several models for road condition monitoring. In total, 59 teams participated in the challenge from 14 different countries. These teams, although provided solutions having better accuracy than the original models included in \cite{maeda2018road}, the new solutions were mostly based on changing the underlying network models and hyper-parameter configurations (\cite{alfarrarjeh2018deep, kluger2018region, wang2018deep, wang2018road}). Although some authors suggested improvements in the dataset but the teams themselves did not modify the dataset in their work.
After that, some researchers focused on experimenting with the underlying dataset by either adding more images to it or by introducing a completely new dataset based on a similar method utilized by \cite{maeda2018road}.
For instance, \cite{angulo2019road} extended this dataset by adding images collected from Italy and Mexico. The size of the dataset was increased to 18034 images, and the authors mostly focused on having a more balanced representation for individual damage classes, especially potholes. 
Similarly, \cite{roberts2020towards} used the data collection app introduced by \cite{maeda2018road} and collected over 7000 road images from Sicily, Italy. Their work utilizes only the newly collected images, but the underlying methodology is similar to \cite{maeda2018road}, except that they also consider severity analysis for the identified road damages. 
The presented study differs from these works in the following ways:
\begin{enumerate}
    \item The Smartphone-based dataset introduced in our work is larger and more heterogeneous, covering multiple countries.
    
    \item These works do not consider the perspective of evaluating the responses of existing or newly proposed models for detecting damages in different countries individually. 
    
    \item The Japanese dataset has been updated by re-annotating and adding more images from Japan, named RDD-2019, in recent work (\cite{maeda2020}). Unlike the previous studies based on the Japanese dataset (2018 version) or models, our work builds upon this updated dataset.

    \item We also analyze the effect of varying the number of images in training dataset on the performance of the trained models. This also lays the foundation for analyzing whether a model trained on data from a single country or multiple countries performs better.
\end{enumerate}

Another work (\cite{du2020pavement}) uses a dataset of 45,788 road images collected from Shanghai and utilize YOLO network for detecting and classifying pavement distresses. However, the images are collected using an industrial high-resolution camera, unlike our work, which is based on the lesser expensive Smartphone-based images.
Similarly, \cite{majidifard2020pavement} use Google street view images considering both top-down as well as wide-view for classification and densification of pavement distresses collected from 22 different pavement sections in the United States. However, the size of the dataset used is limited to only 7237 images. 
Ideally, more than 5000 labeled images are generally required for each class for an image processing based classification task to provide satisfactorily accurate results (\cite{goodfellow2016deep, maeda2020}). This is one of the reasons why there is a need to analyze whether the data already available from another country can be mixed with the local data for improving the performance of a road damage detection and classification model.
This study intends to address the research gaps and provides the recommendations required for practical use of the models by different countries.
The following section provides detail of the proposed dataset and the methodology followed by this study to design the experiments and carry out the requisite analysis. 

\section{Datasets and Methodology}
This section first explains the transition from the dataset RDD-2018 to the proposed dataset, followed by an explanation of the dataset used in this study.

\subsection{RDD-2018}

\begin{table*}
\centering
\caption{Road Damage Types and Definitions considered in \cite{maeda2018road} \label{tab1}}
\label{crackTypeDef}
\scalebox{0.83}{
\begin{tabular}{|ccc|c|c|}
\multicolumn{3}{c}{}                                         &\multicolumn{1}{c}{} &\multicolumn{1}{c}{} \\ \hline
\multicolumn{3}{|c|}{Damage Type}                                           &\multicolumn{1}{|c|}{Detail} &\multicolumn{1}{|c|}{Class Name} \\ \hline \hline
\multicolumn{1}{|c|}{} &  \multicolumn{1}{|c|}{}             & Longitudinal & Wheel-marked part           & D00\\ \cline{4-5}
Crack                  &  \multicolumn{1}{|c|}{Linear Crack} &              & Construction joint part   & D01\\ \cline{3-5}
\multicolumn{1}{|c|}{} &  \multicolumn{1}{|c|}{}             & Lateral      & Equal interval            & D10 \\ \cline{4-5}
\multicolumn{1}{|c|}{} &  \multicolumn{1}{|c|}{}             &              & Construction joint part   & D11\\ \cline{2-5}
                       &  \multicolumn{2}{|c|}{Alligator Crack}             & Partial pavement, overall pavement & D20 \\ \hline
                       &                                     &              & Pothole                   & D40\\ \cline{4-5}
\multicolumn{3}{|c|}{Other Damage}                                      & Cross walk blur           & D43\\ \cline{4-5}
                       &                                     &              & White line blur           & D44\\ \hline
\end{tabular}
}

\end{table*}

RDD-2018 (\cite{maeda2018road}) was introduced and made publicly available in 2018, and since then, it has been used exhaustively by several researchers to propose new methods for automatic condition monitoring of roads as well as other infrastructures. Table \ref{tab1} shows the damage categories and the corresponding definitions used in this dataset.

\subsection{RDD-2019} 
\cite{maeda2020} reviewed and relabelled the contents of the dataset RDD-2018 and added new annotated images to form the new dataset RDD-2019. The number of images with annotations were increased from 9,053 to 13,135. The number of annotations increased from 15,435 to 30,989. 

\subsection{The proposed Road Damage Dataset 2020}

The proposed dataset builds upon the recently introduced RDD-2019 dataset. This is to be noted that the road damage data included in Road Damage Dataset 2018 and 2019 comprise road images collected from only a single country, that is, Japan. However, the dataset proposed in this study considers data collected from multiple countries. The significant points of difference are as follows.

\begin{enumerate}
    \item The total number of images is increased to 26620, almost thrice the prevailing 2018 dataset.
    
    \item New images were collected from India and Czech Republic (partially from Slovakia) to make the data more heterogeneous and train robust algorithms.
   
    \item Unlike previous versions, this dataset considers only four damage categories, comprising mainly of cracks and potholes, namely D00, D10, D20, and D40. Note that the standards related to evaluations of Road Marking deterioration such as Crosswalk or White Line Blur differ significantly across several countries. Thus, these categories were excluded from the study so that generalized models can be trained applicable for monitoring road conditions in more than one country.
    
\end{enumerate}
 
More details related to the newly collected images from India and the Czech Republic are provided as follows.

\subsubsection{Study Area}
Road images for the presented study were collected from India, the Czech Republic and partially, in Slovakia. For the Czech Republic and Slovakia, a large portion of road images was collected in Olomouc, Prague, and Bratislava municipalities and included a mix of 1st class, 2nd class, 3rd class roads, and local roads. A smaller portion of the road image dataset was collected along D1, D2, and D46 motorways to improve the robustness of the trained model. 
Similarly for India, a mixture of images collected from local roads, State Highways, and National Highways was considered, covering the Metropolitan (Delhi, Gurugram) as well as Non-Metropolitan regions (mainly from the state Haryana). All these images were collected from plain regions. Road selection and time of data collection were made based on road accessibility, weather conditions, and traffic volume. 

\subsubsection{Data Collection}
\begin{figure}[ht]
\begin{center}
%\centerline{\includegraphics[width=342pt,height=9pc]{Fig_3_car.png}}
\includegraphics[width=0.25\textwidth]{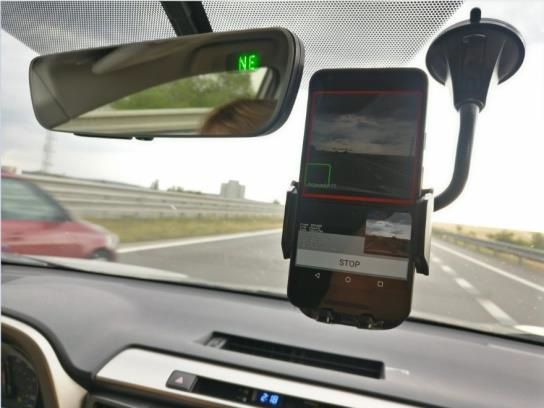}
\caption{Installation Setup of the Smartphone in the Car \label{fig3_car}}
\end{center}
\end{figure}

Road images were captured using a smartphone running a publicly available image-capturing application developed by Sekimoto Lab, The University of Tokyo (\cite{maeda2018road}). The smartphone was installed on the windshield inside the vehicle, and the application captured JPEG images with a resolution of $600 \times 600$ pixels at a rate of one image per second (Figure \ref{fig3_car}). For India, the updated version of the application was used, and the images with a resolution of $720 \times 960$ pixels were captured. These images were resized to $720 \times 720$ to have square images similar to Japan and Czech to maintain uniformity in the dataset for different countries.
This rate was selected to enable the collection of pictures without overlap or leakage when the vehicle traveled at an average speed of approximately 40 km/h (or 25 mph). However, vehicle speed on motorways and some 1st class roads/highways outside the cities was adjusted to comply with the local laws.

For the Czech Republic and Slovakia, a total of 10400 images were collected between March and October 2018 under varying weather and lighting conditions, including sunny, overcast, light rain, and sunset. However, only 3595 images were found suitable for the study. The remaining images which were either blurred or were not covering the required road damage categories or significant portions of the road were discarded. Similarly, for India, a total of 10595 images were captured in October 2019, out of which, 9892 images were finally selected. 

\subsubsection{Data Classification/Damage Categories}
Based on the Japanese Maintenance Guidebook for Road Pavement (JRA, 2013), \cite{maeda2018road} use eight damage categories in total. This classification covers two types of deterioration, namely pavement deterioration (D00, D01, D10, D11, D20, D40) and road marking deterioration (D43, D44), as shown in Table \ref{tab1}.
However, the standards for evaluating the road pavement conditions are not the same across the countries and vary significantly. For instance, the Czech Catalog of Deficiencies in Flexible Pavements, Czech Ministry of Transport(2009) recognizes 28 different pavement deterioration divided into three main categories, including loss of asphalt mixture, cracks, and deformations. This catalog does not recognize Crosswalk marking deterioration as a type of pavement deficiency.
Similarly, for India, the code \cite{IRC82_2015} provides guidelines for the maintenance of bituminous/flexible pavement in India. It identifies four main types of pavement distresses, namely, Cracking, Deformation (rutting, upheavals, etc.), Disintegration (raveling, potholes, edge breaking, etc.), and Surface Defects that relate to inapt quality and quantity of bitumen. 
Since the aim of this study is to consider the perspective of monitoring road conditions in more than one country, we include only the four major categories of damage mainly comprising cracks and potholes, which are generally common to almost all the countries. 
Thus, for this study, the damage categories considered are - D00 to represent Longitudinal/ Parallel cracks, D10 for Transverse/Perpendicular cracks, D20 for Alligator/Complex cracks, and D40 for Potholes. Some other categories which are indistinguishable in the images from the above mentioned four cases, for instance, rutting is marked as D40 in Japanese dataset (\cite{maeda2018road}), are covered along with these so as to provide a generalized model. 
Further, this is to be noted that the existing Japanese datasets (RDD-2018 and 2019) also contain annotations for damage categories (D01, D11, D43, D44, D50), which are not included in this study. Our work makes no changes to these already available annotated images. Instead, it uses algorithms that read only the annotations related to the desired four categories.

\subsection{Data Annotation}

\begin{figure}[ht]
%\centerline{\includegraphics[width=342pt,height=9pc]{Fig_4_annotation.png}}
\centerline{\includegraphics[width=0.5\textwidth]{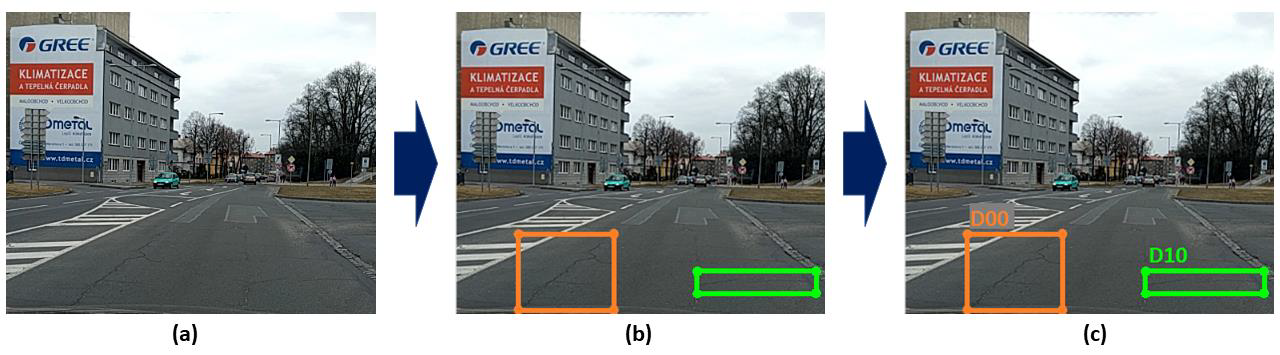}}
\caption{Annotation Pipeline: (a) original image, (b) image with bounding boxes, (c) final annotated image containing bounding boxes and class labels \label{fig4}}
\end{figure}

Data annotation is a fundamental part of what makes many machine learning projects function accurately. It provides the initial setup for teaching a deep neural network to recognize objects and discriminate them against various input images. In this work, the collected road images were manually annotated using labelImg software. The annotation pipeline is presented in Figure \ref{fig4}. Annotation is a labor-intensive procedure and took considerable time because the complete dataset was first manually reviewed on the screen and then all recognized deficiencies were annotated by enclosing them with bounding boxes and classified by attaching the proper class label. Class labels and bounding box coordinates, defined by four decimal numbers $(x_{min}, y_{min}, x_{max}, y_{max})$, were stored in the XML format similar to PASCAL VOC (\cite{everingham2010pascal}). Next, the data was converted into TFRecord file format, as required by the TensorFlow Object Detection API (\cite{Intel2018}).

\subsubsection{Statistics}

\begin{figure*}[ht]
\centerline{\includegraphics[width=342pt,height=19pc]{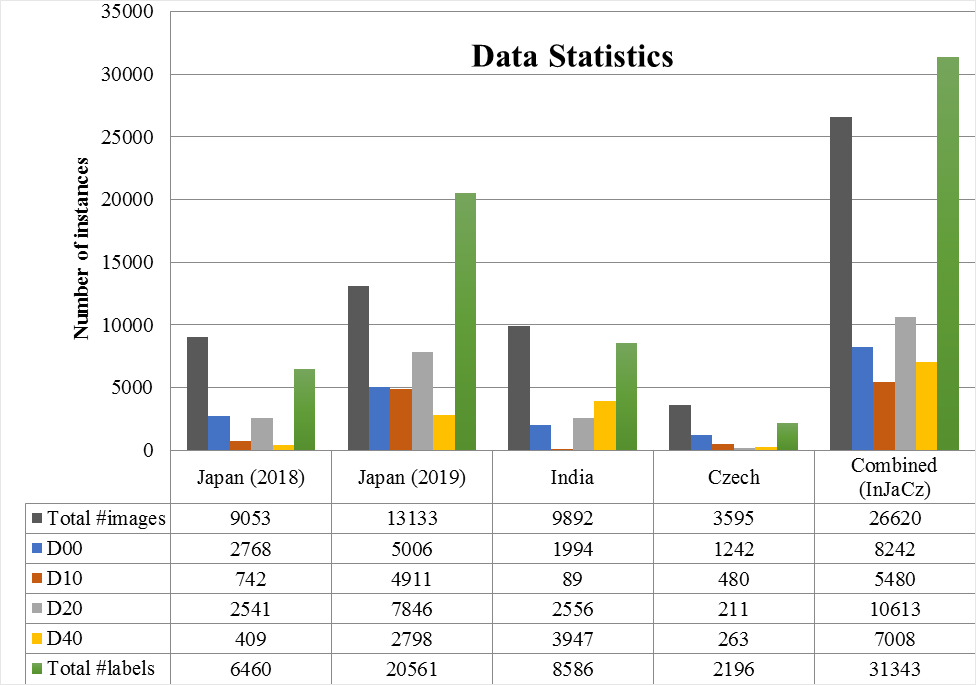}}
\caption{Statistics for the number of damage instances included in the underlying datasets \label{fig5}}
\end{figure*}

Figure \ref{fig5} shows the statistics for the proposed dataset. A comparatively lesser number of images are used from the countries India and Czech than what was available from Japan in RDD-2019. This is because our motive is to analyze and provide solutions in which the dataset already available from other countries can be utilized to design models for new countries, such that the requirement for collecting new local data is minimized.
Further, this can be noted that the number of instances for different damage categories in Indian and Czech datasets is a bit dis-balanced. This leads to the requirement of analyzing if the data from another country can be used to create a balanced representation. 
For example, only 89 instances of transverse cracks, that is, D10 are present in the 9892 images available from India, which is very less. Combining it with dataset available from Japan may help in representing this class better while training the requisite neural network models. Similarly, for alligator cracks, D20, and potholes, D40, in the case of Czech data, adding data from another country or collecting new local data becomes a necessity for training efficient models.
The following section illustrates the methodology followed in this study.

\subsection{Experimental Setup and the Methodology}

\begin{table}[t]
\centering
\caption{\label{t2}Naming Convention for the Models}
\label{ModelNames}

\scalebox{0.85}{
%\begin{tabular}{ccccc}
\begin{tabular}{|c|c|c|c|c|}
\hline
\multirow{2}{*}{\textbf{Model Name}} & \multicolumn{3}{c}{\textbf{Training Data (\#images)}} & \multirow{2}{*}{\textbf{Tested for}} \\ \cline{2-4}
 & \textbf{Japan} & \textbf{India} & \textbf{Czech} &  \\ \hline
Ja\_2k & 2000 &  &  & \multirow{5}{*}{Japan, India, Czech} \\ %\cline{1-4}
Ja\_4k & 4000 &  &  &  \\ %\cline{1-4}
Ja\_6k & 6000 &  &  &  \\ %\cline{1-4}
Ja\_8k & 8000 &  &  &  \\ %\cline{1-4}
Ja\_10k & 10000 &  &  &  \\ \hline
In\_2k &  & 2000 &  & \multirow{4}{*}{India} \\ %\cline{1-4}
In\_4k &  & 4000 &  &  \\ %\cline{1-4}
In\_6k &  & 6000 &  &  \\ %\cline{1-4}
In\_8k &  & 8000 &  &  \\ \hline
Cz\_2k &  &  & 2000 & Czech \\ \hline
InJa\_12k & 10000 & 2000 &  & \multirow{4}{*}{India \& Japan} \\ %\cline{1-4}
InJa\_14k & 10000 & 4000 &  &  \\ %\cline{1-4}
InJa\_16k & 10000 & 6000 &  &  \\ %\cline{1-4}
InJa\_18k & 10000 & 8000 &  &  \\ \hline
JaCz\_11k & 10000 &  & 1000 & \multirow{2}{*}{Czech} \\ %\cline{1-4}
JaCz\_12k & 10000 &  & 2000 &  \\ \hline

\end{tabular}
}
\end{table}

The study involves training and evaluating 16 deep neural network models considering 30 scenarios based on different combinations of the test and train datasets, as described in table \ref{ModelNames}. The critical differences in the models are: 

\begin{enumerate}[]
    \item Source of the training data: Following two cases were considered -
    
    \begin{enumerate}[]
        \item Single Source Models – It covers the models trained on a single source of data (Japan) applied to test data from different countries (India, Japan, Czech).
        
        \item 	Multiple Source Models – It covers the following two categories:
        
        \begin{enumerate}[]
            \item Pure Modelling or Same Source, Same Target Modelling: A model trained and tested on data collected from the same country.
            
            \item Mixed Modelling: It includes mixing the local data of some other country, generally the target country, with publicly available data from Japan for training the models.
        \end{enumerate}
      
    \end{enumerate}
    \item  Size of the training data: The number of images considered for training the models was varied from 2000 to 18000 based on the availability of the data from different countries. This was done to analyze the effect of varying the size of training data on the performance of the models. 
\end{enumerate}

Further, since the images available from different sources were originally in different resolutions as described in data collection, images were re-scaled to a resolution of $300 \times 300$ pixels while training the models.
TensorFlow API $1.80$ and official Python distribution 3.6.5 were used to construct, train, and deploy the object detection models based on transfer learning using MobileNet. The learning rate for training the models was set to $0.003$.

\section{Results and Analysis}
\subsection{Evaluation Parameters}
The performance of the trained model was estimated by using three indices, namely, precision, recall, and F1-score. Precision indicates the percentage of correctly predicted features (i.e., true positives) out of the total number of predicted features (i.e., true and false positives). The recall shows the percentage of correctly predicted features out of the total number of features present in the actual class (i.e., true positives and false negatives). 

\begin{figure}[ht]
\centerline{\includegraphics[width=0.51\textwidth{}]{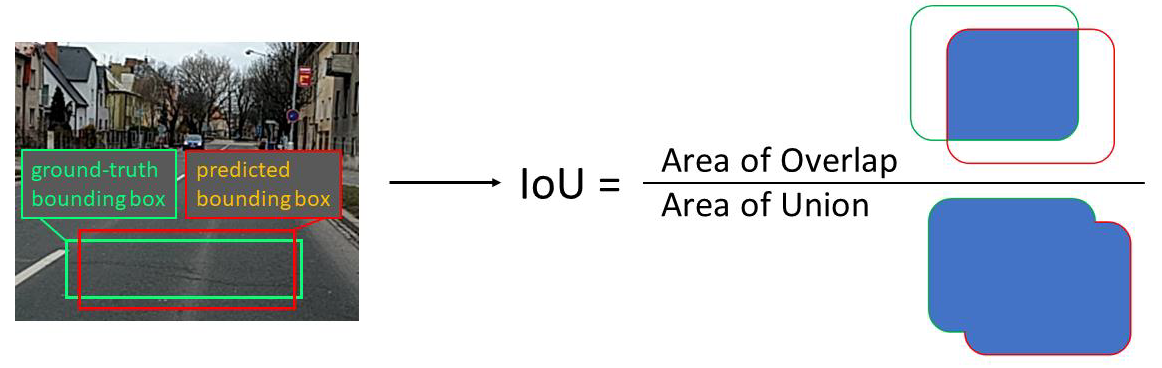}}
\caption{An illustration for calculating Intersection over Union (IoU) \label{fig6}}
\end{figure}

Precision and recall are both based on evaluating Intersection over Union (IoU), which is defined as the division of the area overlap between predicted and ground-truth bounding boxes by the area of their union, as shown in Figure \ref{fig6} . Because precision and recall counter each other and increasing one of them will usually reduce the other, The most common way to reach a balance between these metrics is to use F1-score. It measures the overall model’s accuracy and is calculated as:

\begin{equation}\label{eq1}
F_1 = 2 \times {\frac {(precision \times recall)}{(precision+recall)}}
\end{equation}
Maximizing the F1-score ensures reasonably high precision and recall.

For evaluation of the model’s performance, the IoU threshold was set to $0.5$, which represents an evaluation metric prescribed by the PASCAL VOC object detection competition (\cite{everingham2010pascal}).

\subsection{Evaluation results}

\begin{table*}[ht]
\centering
\caption{\label{t3_wide}F1-Score for the experiments}
\label{F1_Score_table}
\scalebox{0.93}{
\begin{tabular}{|c|c|c|c|c|c|c|}
\hline
 &  &  & \multicolumn{4}{c|}{\textbf{Damage Category}} \\ \cline{4-7} 
\multirow{-2}{*}{\textbf{Experiment}} & \multirow{-2}{*}{\textbf{Model Name}} & \multirow{-2}{*}{\textbf{Test Data(1000 images)}} & \textbf{D00} & \textbf{D10} & \textbf{D20} & \textbf{D40} \\ \hline
\rowcolor[HTML]{F2F2F2} 
E-1 & Ja\_2k & \cellcolor[HTML]{F2F2F2} & 0.196 & 0.084 & 0.404 & 0.305 \\ \cline{1-2} \cline{4-7} 
\rowcolor[HTML]{F2F2F2} 
E-2 & Ja\_4k & \cellcolor[HTML]{F2F2F2} & 0.257 & 0.138 & 0.450 & 0.305 \\ \cline{1-2} \cline{4-7} 
\rowcolor[HTML]{F2F2F2} 
E-3 & Ja\_6k & \cellcolor[HTML]{F2F2F2} & 0.268 & 0.154 & 0.421 & 0.356 \\ \cline{1-2} \cline{4-7} 
\rowcolor[HTML]{F2F2F2} 
E-4 & Ja\_8k & \cellcolor[HTML]{F2F2F2} & 0.293 & 0.194 & 0.435 & 0.334 \\ \cline{1-2} \cline{4-7} 
\rowcolor[HTML]{F2F2F2} 
E-5 & Ja\_10k & \multirow{-5}{*}{\cellcolor[HTML]{F2F2F2}Japan} & 0.308 & \textbf{0.228} & 0.504 & 0.401 \\ \hline
\rowcolor[HTML]{FFF2CC} 
E-6 & Ja\_2k & \cellcolor[HTML]{FFF2CC} & 0.024 & nan & 0.050 & 0.015 \\ \cline{1-2} \cline{4-7} 
\rowcolor[HTML]{FFF2CC} 
E-7 & Ja\_4k & \cellcolor[HTML]{FFF2CC} & nan & nan & 0.022 & 0.048 \\ \cline{1-2} \cline{4-7} 
\rowcolor[HTML]{FFF2CC} 
E-8 & Ja\_6k & \cellcolor[HTML]{FFF2CC} & 0.030 & nan & 0.034 & 0.010 \\ \cline{1-2} \cline{4-7} 
\rowcolor[HTML]{FFF2CC} 
E-9 & Ja\_8k & \cellcolor[HTML]{FFF2CC} & 0.032 & nan & 0.022 & 0.052 \\ \cline{1-2} \cline{4-7} 
\rowcolor[HTML]{FFF2CC} 
E-10 & Ja\_10k & \multirow{-5}{*}{\cellcolor[HTML]{FFF2CC}India} & 0.051 & nan & 0.028 & 0.036 \\ \hline
\rowcolor[HTML]{E2EFD9} 
E-11 & Ja\_2k & \cellcolor[HTML]{E2EFD9} & 0.110 & 0.062 & 0.064 & Nan \\ \cline{1-2} \cline{4-7} 
\rowcolor[HTML]{E2EFD9} 
E-12 & Ja\_4k & \cellcolor[HTML]{E2EFD9} & 0.156 & 0.082 & 0.019 & Nan \\ \cline{1-2} \cline{4-7} 
\rowcolor[HTML]{E2EFD9} 
E-13 & Ja\_6k & \cellcolor[HTML]{E2EFD9} & 0.158 & 0.022 & 0.046 & Nan \\ \cline{1-2} \cline{4-7} 
\rowcolor[HTML]{E2EFD9} 
E-14 & Ja\_8k & \cellcolor[HTML]{E2EFD9} & 0.120 & 0.109 & 0.045 & Nan \\ \cline{1-2} \cline{4-7} 
\rowcolor[HTML]{E2EFD9} 
E-15 & Ja\_10k & \multirow{-5}{*}{\cellcolor[HTML]{E2EFD9}Czech} & 0.175 & 0.097 & 0.022 & Nan \\ \hline
\rowcolor[HTML]{FFF2CC} 
E-16 & In\_2k & \cellcolor[HTML]{FFF2CC} & 0.141 & nan & 0.432 & 0.266 \\ \cline{1-2} \cline{4-7} 
\rowcolor[HTML]{FFF2CC} 
E-17 & In\_4k & \cellcolor[HTML]{FFF2CC} & 0.140 & nan & 0.448 & 0.295 \\ \cline{1-2} \cline{4-7} 
\rowcolor[HTML]{FFF2CC} 
E-18 & In\_6k & \cellcolor[HTML]{FFF2CC} & 0.199 & nan & \textbf{0.502} & \textbf{0.333} \\ \cline{1-2} \cline{4-7} 
\rowcolor[HTML]{FFF2CC} 
E-19 & In\_8k & \multirow{-4}{*}{\cellcolor[HTML]{FFF2CC}India} & \textbf{0.226} & nan & 0.442 & 0.307 \\ \hline
\rowcolor[HTML]{E2EFD9} 
E-20 & Cz\_2k & Czech & 0.263 & 0.071 & \textbf{0.282} & \textbf{0.196} \\ \hline
\rowcolor[HTML]{FFF2CC} 
E-21 & InJa\_12k & \cellcolor[HTML]{FFF2CC} & 0.159 & nan & 0.325 & 0.276 \\ \cline{1-2} \cline{4-7} 
\rowcolor[HTML]{FFF2CC} 
E-22 & InJa\_14k & \cellcolor[HTML]{FFF2CC} & 0.129 & nan & 0.391 & 0.257 \\ \cline{1-2} \cline{4-7} 
\rowcolor[HTML]{FFF2CC} 
E-23 & InJa\_16k & \cellcolor[HTML]{FFF2CC} & 0.200 & nan & 0.403 & 0.311 \\ \cline{1-2} \cline{4-7} 
\rowcolor[HTML]{FFF2CC} 
E-24 & InJa\_18k & \multirow{-4}{*}{\cellcolor[HTML]{FFF2CC}India} & 0.224 & nan & 0.487 & 0.305 \\ \hline
\rowcolor[HTML]{F2F2F2} 
E-25 & InJa\_12k & \cellcolor[HTML]{F2F2F2} & 0.327 & 0.174 & 0.522 & 0.369 \\ \cline{1-2} \cline{4-7} 
\rowcolor[HTML]{F2F2F2} 
E-26 & InJa\_14k & \cellcolor[HTML]{F2F2F2} & \textbf{0.361} & 0.177 & 0.502 & \textbf{0.434} \\ \cline{1-2} \cline{4-7} 
\rowcolor[HTML]{F2F2F2} 
E-27 & InJa\_16k & \cellcolor[HTML]{F2F2F2} & 0.343 & 0.111 & 0.512 & 0.366 \\ \cline{1-2} \cline{4-7} 
\rowcolor[HTML]{F2F2F2} 
E-28 & InJa\_18k & \multirow{-4}{*}{\cellcolor[HTML]{F2F2F2}Japan} & 0.335 & 0.128 & \textbf{0.524} & 0.336 \\ \hline
\rowcolor[HTML]{E2EFD9} 
E-29 & JaCz\_11k & \cellcolor[HTML]{E2EFD9} & \textbf{0.324} & \textbf{0.177} & 0.204 & 0.076 \\ \cline{1-2} \cline{4-7} 
\rowcolor[HTML]{E2EFD9} 
E-30 & JaCz\_12k & \multirow{-2}{*}{\cellcolor[HTML]{E2EFD9}Czech} & 0.318 & 0.128 & 0.228 & 0.092 \\ \hline
\end{tabular}
}
\end{table*}

Table \ref{F1_Score_table} presents the value of the F1-score obtained for different damage categories in several experiments, respectively. The highlighted entries show the maximum F1-score obtained for each country for different damage categories.

\subsection{Empirical Analysis}

\subsubsection{Single Source Modeling}

\begin{figure}[ht]
\centerline{\includegraphics[width=0.5\textwidth]{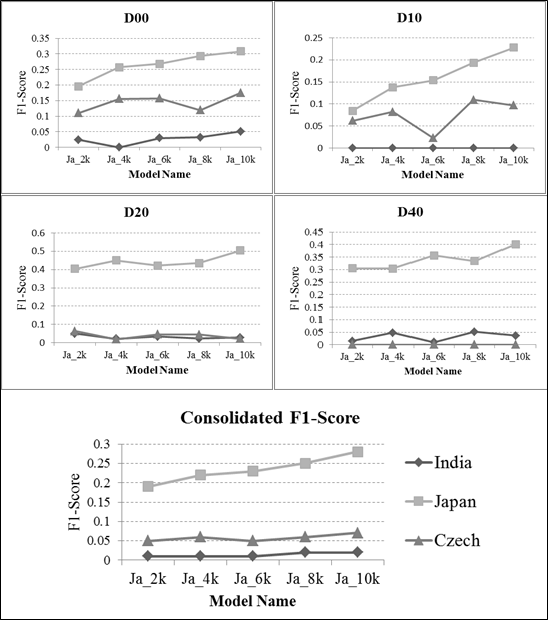}}
\caption{Damage category-wise and Consolidated F1-Score for Models trained on Japanese data applied to test data from different countries.\label{fig7}}
\end{figure}

Figure \ref{fig7} shows the results for different damage categories when a model trained on Japanese data is applied to different target countries by using test data constituting 1000 images from Japan, India, and Czech, respectively.

As depicted by the individual and the consolidated chart, the model trained on Japanese data, though worked fine for Japanese test data; the performance is abysmal when applied to test data from India and Czech, irrespective of the number of images used for training. Though, the model performs better for some damage categories viz. a viz. others, the overall performance is below acceptable limits.

\subsubsection{Multiple Source Modeling}
The study analyzes the models trained using multiple sources of data based on the target or the test data, as illustrated below.

\paragraph{Target: Japan}
\begin{figure}[ht]
\centerline{\includegraphics[width=0.5\textwidth]{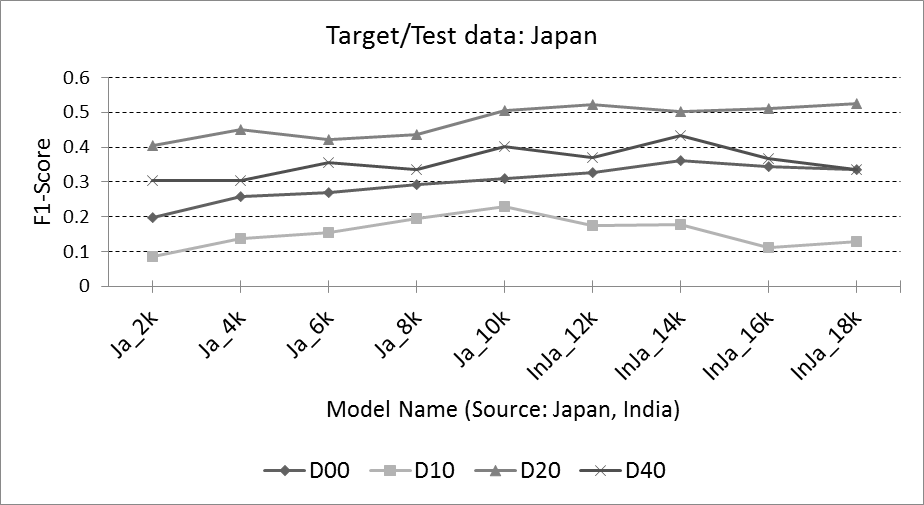}}
\caption{Model name vs F1-Score (Target: Japan) \label{fig8}}
\end{figure}

Figure \ref{fig8} shows the results for various models applied to Japanese test data. Clearly, the models having training data as a mixture of images from India and Japan show better performance for Japan for D00, D20, and D40.
For D10, the performance shows a downfall on increasing the number of images added from India because the number of instances for the D10 category is almost negligible in Indian data (89 in total). And, thus, as the total number of images increases on adding Indian data to Japanese data, the proportion of D10 instances in the total number of damages decreases, leading to the decline in the performance of the final trained model in identifying D10.

\paragraph{Target: India} 

\begin{figure}
\centerline{\includegraphics[width=0.5\textwidth]{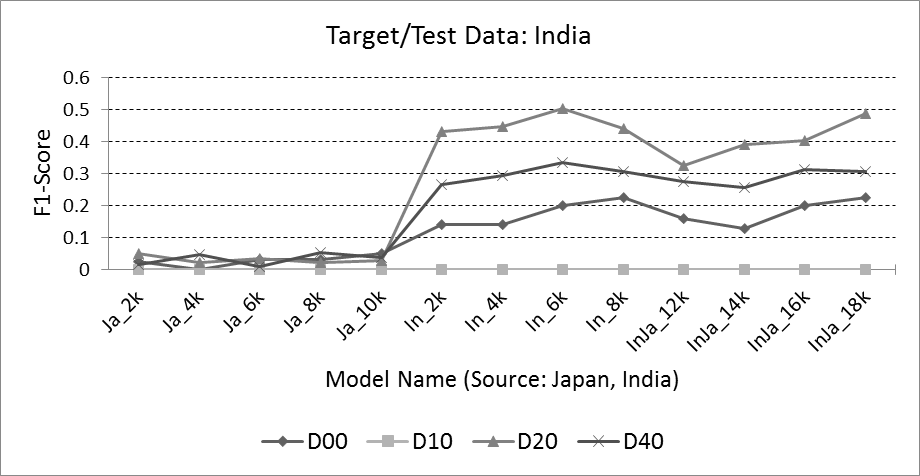}}
\caption{Model name vs F1-Score (Target: India) \label{fig9}}
\end{figure}

Figure \ref{fig9} shows the compiled results for different models applied to test data from India. The results illustrate that a model trained using Indian data (either pure or mixed in the form of InJa models) perform significantly better than the model trained using only Japanese data. 
Notably, even the models trained using Indian data fail to perform satisfactorily for Transverse Cracks (D10) in Indian data. This is because only 89 instances of transverse cracks are present in the 9892 images available from India, leading to the presence of only 9 to 10 images in Indian test data constituting 1000 images in total. These 10 images are not sufficient for evaluating the performance of the object detection models leading to nan(Not a number) values in F1-score. 
For other categories, the observed trend show adding images from Japan does not help much in improving the performance of the models trained using only Indian data. Nonetheless, the performance of pure Indian models is slightly better than the mixed InJa models, despite having smaller training datasets.

\begin{figure}
\centerline{\includegraphics[width=0.5\textwidth]{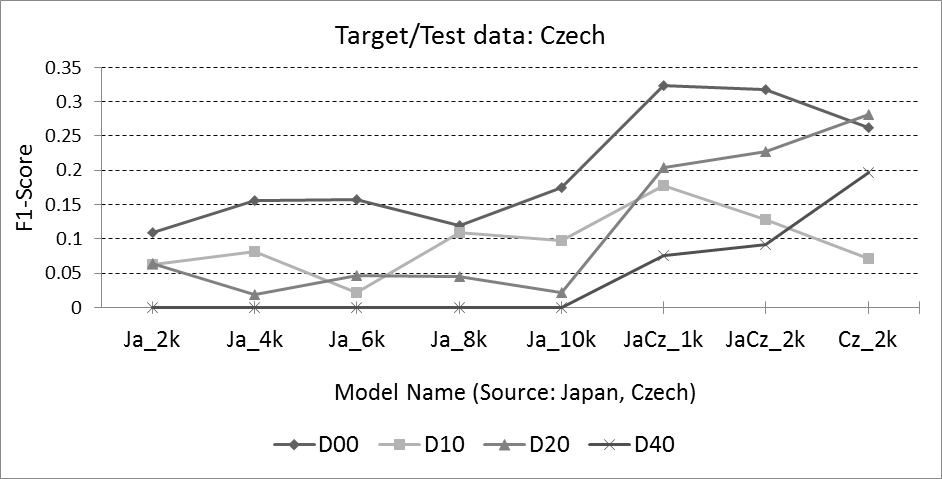}}
\caption{Model name vs F1-Score (Target: Czech) \label{fig10}}
\end{figure}

\paragraph{Target: Czech}
Figure \ref{fig10} shows the results for different models applied to the test data from the Czech. Analyzing the graph from right to left shows, a model trained using only 2000 images of Czech (Cz\_2k) shows better average performance than the model trained using Japanese data constituting up to 10000 images. However, this may be a case of overfiting.
Further, mixing the images from Czech with the already available Japanese data (10000 images) shows some improvement over the case when using data from a single country. Two cases were considered for the analysis, based on the number of images used from Czech. The observations are listed as follows: 

\begin{enumerate}
    \item Adding images from Japan to Czech data helps in improving the performance for Linear Cracks, both longitudinal as well as transverse.
    \item For Alligator cracks and Potholes – the pure Czech model seems to outperform the mixed JaCz models. This can be explained as follows. The number of instances for D20 and D40 in Czech data is very less, and mixing a huge amount of Japanese data for training the model decreases the proportion of Czech D20 and D40 instances further. This leads to the dominated effect of corresponding features learned from Japanese data and ultimately to degraded performance of JaCz models for identifying D20 and D40 of Czech.
\end{enumerate}

\subsection{Visual Analysis}

\begin{figure*}[ht]
\centerline{\includegraphics[width=1\textwidth]{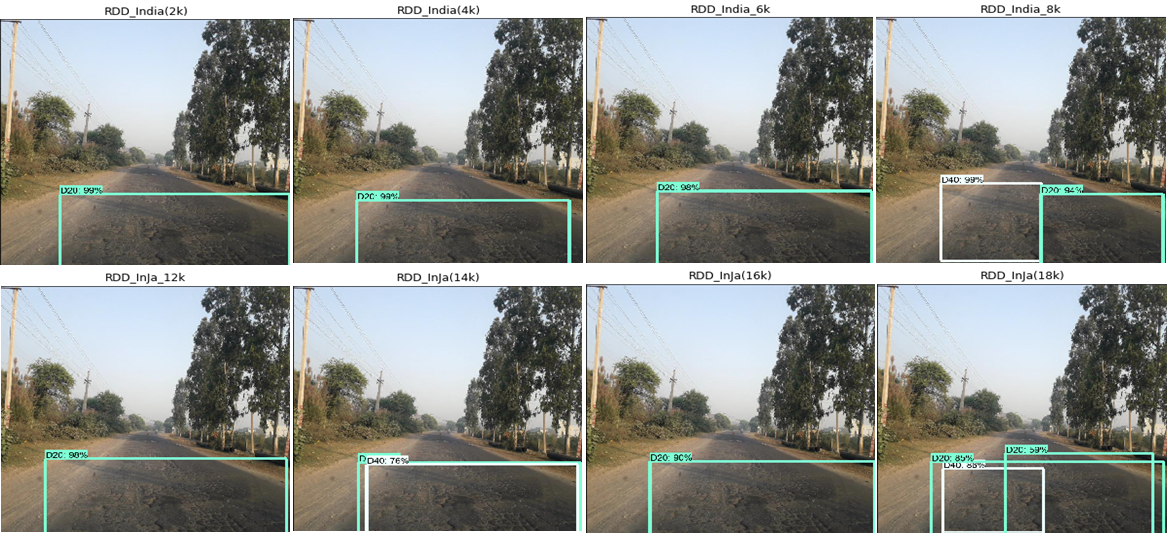}}
\caption{Labels predicted for Sample Image 1 by models trained using RDD\_India and RDD\_InJa \label{fig11a}}
\end{figure*}

\begin{figure*}[ht]
\centerline{\includegraphics[width=1\textwidth]{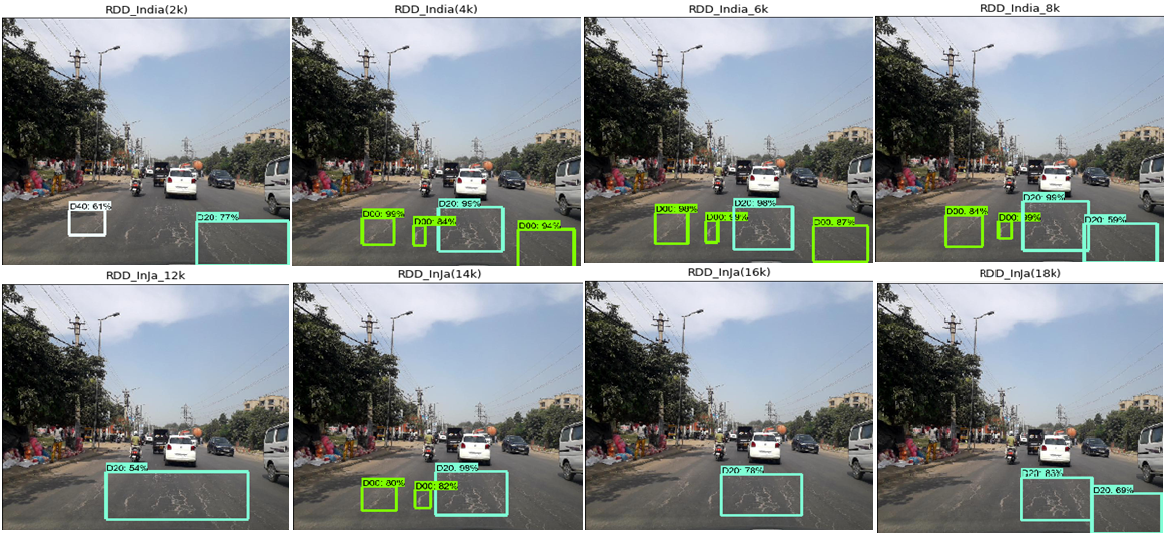}}
\caption{Labels predicted for Sample Image 2 by models trained using RDD\_India and RDD\_InJa \label{fig11b}}
\end{figure*}

As described in the empirical analysis, the behavior of models trained for Japan or Czech is easily understandable. However, the models trained using only the Indian datasets show comparable performance to the models trained using the Indian dataset mixed with 10000 Japanese images, which is a bit unusual. Usually, the performance should get improved on increasing the size of the datasets by such a large extent. To better understand this uncommon behavior of the models, we analyzed the corresponding prediction results visually. Figure \ref{fig11a} and \ref{fig11b} show the labels predicted by several models trained using Indian (In\_2k, In\_4k, In\_6k, In\_8k) and combined Indian-Japanese (InJa\_12k, InJa\_14k, InJa\_16k, InJa\_18k) datasets for different Indian road images. 

The image in figure \ref{fig11a} shows a high severity alligator crack intermixed with pothole-like structures. For the given image, India\_8k identified that the alligator cracks have taken the form of potholes. InJa\_14k could identify the presence of D40 with D20, though not individually. InJa\_18k provided the most precise result. 

Clearly, there cannot be a single correct label boundary for this type of image. The labels predicted by all the models are acceptable when a comprehensive survey for monitoring road conditions is required. However, since the evaluation of object detection models involves comparing the predicted labels with the information in the ground truth file, some of the predicted labels, although acceptable, get marked as false examples. When many such images are present in the dataset, it results in low performance of some of the models. This may be a reason for the low accuracy of InJa viz.a.viz. Indian models, despite having larger training datasets. The features learned by InJa models are a bit dominated by the Japanese dataset since the number of images used from Japan is more than that of India in all the cases. 

For the image in figure \ref{fig11b}, the outputs of Indian models are too much specific, providing a hint that these models may be overfitting the Indian data. For InJa models, InJa\_14k shows better accuracy than the remaining models.
To analyze if the Indian models are overfitting the Indian data, we analyzed the behavior of these models for data collected from Japan and Czech.

\begin{figure}[ht]
\centerline{\includegraphics[width=0.5\textwidth]{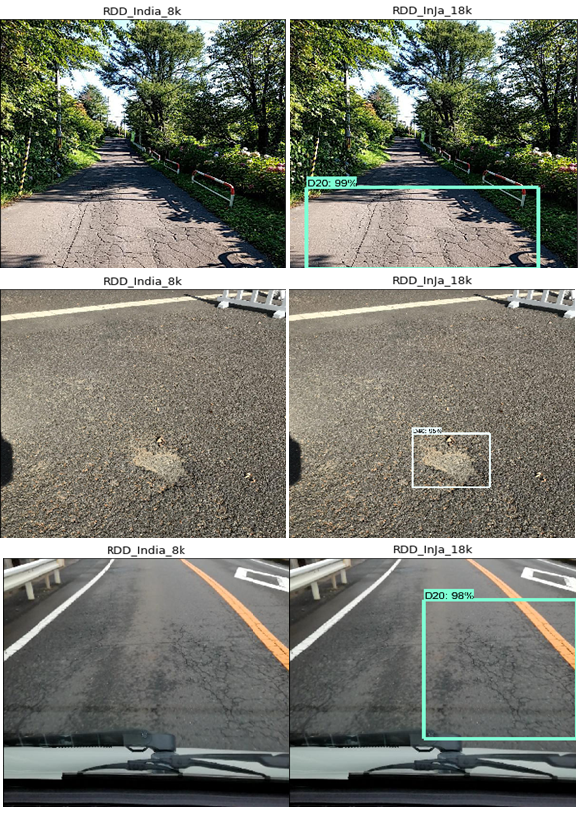}}
\caption{Sample images from Japan and Czech for comparing the performances of In\_8k and InJa\_18k models (Cases of missed detection by In\_8k) \label{fig12}} 
\end{figure}

\begin{figure}[ht]
\centerline{\includegraphics[width=0.5\textwidth]{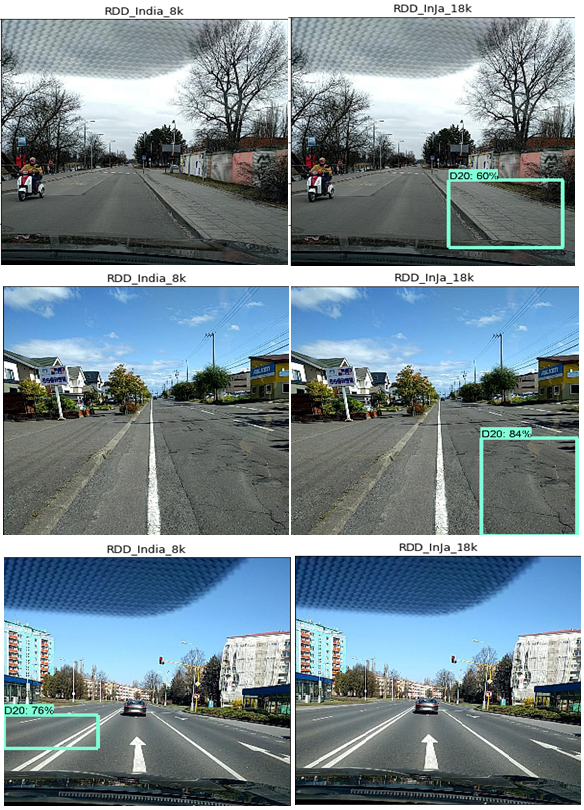}}
\caption{Sample images from Japan and Czech for comparing the performances of In\_8k and InJa\_18k models (Cases of false detection) \label{fig13}} 
\end{figure}

Figure \ref{fig12} shows some sample images from Japan and Czech in which In\_8k missed some damages, but InJa\_18k identified them correctly. Figure \ref{fig13} shows the images for false detection by Indian or InJa models. From top to bottom, the first image demonstrates the falsely detected D20 on the roadside by the InJa model. However, the pattern is relatable to D20. For the second image, the pothole-like structures were missed entirely by In\_8k and mis-classified by InJa\_18k. Thus, in case the motive of the survey is to detect the presence or absence of the damages, the performance of InJa\_18k can be counted as better than In\_8k. The last image shows the case of false-positive detection by India\_8k. 

In summary, it can be said that although the models trained using only Indian data show performance comparable to models trained by mixing Japanese data with Indian datasets (InJa), InJa models are more generalized. Also, for InJa models, InJa\_14k is found to have the best overall performance, combining the evaluations for both India and Japan, and can be used efficiently for detecting and classifying road damages in both the countries.

\section{Discussion}
Road maintenance plays an important role in the socioeconomic development of a country. It requires regular assessment of road conditions, which is usually carried out individually by several state agencies. Some agencies perform pavement condition surveys using road survey vehicles equipped with a multitude of sensors to evaluate pavement conditions and deterioration. In these vehicles, laser line-scan cameras and three-dimensional($3D$) cameras are generally employed to capture pavement surfaces with the best possible quality and resolution. However, such imaging equipment mounted on dedicated vehicles is expensive and is often un-affordable for local agencies with limited budgets. This leads to the requirement of low-cost methods capable of comprehensively surveying road surfaces, such as methods that can be implemented using Smartphones. The municipalities in Japan have already started using such Smartphone-based applications; however, many other countries still lack proper research in this direction. This work is an attempt to systematically analyze the opportunities available for these countries when one other country has developed such an application and has made the underlying data and models available publicly.
The study involves training 16 models evaluated using 30 experiments based on the proportion of data used from different countries. 
Notably, the performance of the model trained using single-source of data, in this case, Japan, varies for different targets significantly, indicating that the same model cannot be used everywhere. For multiple source models (pure vs. mixed modeling), the performance varies based on the target country for which the model is trained. 
For Japan, adding images from India to the Japanese dataset helps in improving the performance for longitudinal cracks, alligator cracks as well as potholes. For transverse cracks, the performance was degraded on adding Indian data due to an overall decrease in the proportion of these cracks in total data when Japanese data is mixed with Indian data with an extremely low number of transverse cracks ($<1\%$). 

For India, adding images from Japan to Indian data, though, did not show much improvement in performance with respect to empirical analysis, but improves the generalization capability of the model as demonstrated by the qualitative analysis. 
For Czech, adding images from Japan helps in improving the performance for Linear Cracks. For Alligator cracks and Potholes – the no. of instances in Czech data are very less, and increasing the proportion of Japanese data for training the model leads to the dominated effect of corresponding features learned from Japanese data and ultimately to degraded performance for Czech.

Overall, it can be concluded that mixed source modeling outperforms single-source as well as pure modeling in the case of multiple sources. Further, the following observations can be made:

\begin{enumerate}
    \item Adding data from other countries helps in increasing the generalizability of the models, as demonstrated by the performance of several Indian, Japanese, and mixed India-Japan models.
    \item The effect on improving the performance by adding data from another country may not be the same/bidirectional for two countries. For instance, mixing Indian and Japanese data helped in improving the performance for Japanese roads; however, for Indian roads, only the generalizability of the models was increased.
    \item Visual or qualitative analysis is required for accurately assessing the performance across the countries. For example: For mixed damages (like the cases when potholes and Alligator Cracks are inter-mixed for India), there can be multiple possible predictions which are correct, but are marked as incorrect due to mismatching with the information listed in Ground-Truth xmls. Qualitative analysis helps in assessing those cases.
\end{enumerate}

The following subsections present the proposed recommendations and discuss the scope for extending the work in the future.

\subsection{Recommendations}
Based on the results and analysis, the study proposes the following recommendations for readers of other countries when one country, (Japan, in this case) releases its data and model:
\begin{enumerate}
    \item Can they utilize the Japanese model? Would the performance of the model inherited from another country when applied to their country remain the same?
        \begin{itemize}
            \item Mostly the performance of a model significantly degrades when applied to road images from another country (as evident from our experiments).
            
            \item Readers should perform qualitative analysis by testing the Japanese model on a few images from their country. If the performance is not satisfactory, the readers can follow the steps provided in our work and get a new model for their country using the Japanese model as the baseline.
        \end{itemize}
    
    \item Should they collect images from their country as well? If yes, what effect would these images have on the performance?
        \begin{itemize}
            \item Yes. The local images help in improving the performance of the model trained on foreign data.
        \end{itemize}
    
    \item Should they mix the images from two countries to train the model or use only the images from their country (pure modeling vs. mixed modeling)?
        \begin{itemize}
            
            \item Our research recommends the mixing of images collected from different countries due to the following reasons: 
            
            \begin{enumerate}
            
            \item The models can automatically detect and classify road damages quickly once trained using the appropriate data; however, collecting the road images and preparing the corresponding annotations for Ground-Truth information is a time-taking task. Mixing the images already available from other countries helps in increasing the size of the dataset in lesser time.
            
            \item It prevents overfitting and improves the generalizability of the model.
            
            \item Some damage classes can be better represented using data from other countries (For example, D10 for India).
            
            \item The proposed models are based on transfer learning, and usually, the available models used as a baseline for transfer learning are trained on any general dataset, not necessarily related to road damages. If the road damage data or model is available even though from some other country, using it as the baseline helps the new model learn the required features for automatic road damage detection, faster.
            
        \end{enumerate}
        \end{itemize}
\end{enumerate}

\subsection{Future Scope}

\begin{enumerate}
    \item This research explores the ways to utilize the data and models available from one country to design road damage detection and classification models for other countries. In the future, the same prototype can be extended to propose a single standardized model that is applicable globally or at least to a set of countries having identical road conditions. 
    
    \item Further, only the relative performance of models trained using different datasets were required to carry out this study. Future work can focus on the actual performance of the models leading to designing models with improved accuracy on a similar line.
    
    \item Furthermore, this work can be used as a baseline, and the experiments can be repeated by collecting more images from different countries, including India, Japan, and Czech, under different seasonal conditions for better representing each damage class and improving the robustness of the overall detection system for all damage categories.
    
    \item An additional increase in coverage and turnaround time could be achieved by installing the road damage detection system on smartphones, and car vehicle recorders mounted in the vehicles operated by municipalities, such as public transport or waste collection vehicles.
    
    \item Finally, public participation through the release of a free smartphone application used by the public for uploading geo-tagged images of poorly performing road sections could provide additional data on the current state of the local road network in several countries.
\end{enumerate}

\section{Conclusion}
The presented study proposes a dataset of 26620 road images collected from Japan, India, and the Czech Republic, and analyzes the methods that implement deep learning for automated detection and classification of road damage. These methods provide a base for quickly and cost-effectively surveying the road network by mounting smartphones in the vehicles. A smartphone application based on such methods has already been designed for Japan and is being used by several Japanese municipalities for efficient road condition monitoring since 2018. However, several other countries still lack proper research in this direction. Whether these countries can directly import Japanese or some other country’s models and application or do they need to make any changes, for monitoring conditions of their local roads, is still unaddressed. This research provides recommendations for road agencies of such countries by demonstrating 30 experiments based on different combinations of data collected from Japan, India, and Czech. The main conclusion is that the countries can mix the already available Japanese data with their local road data and can design an efficient model of their own on a similar line as done in this work.
Additionally, the study proposes Japanese data-based models that can efficiently detect and classify road damages in India and the Czech Republic. Thus, the study lays down a foundation for designing a globally applicable standardized model for road damage detection and classification and is useful for pavement engineers, road authorities, municipalities, and researchers from several countries.

% The authors declare no potential conflict of interests.

\balance
%\nocite{*}% Show all bib entries - both cited and uncited; comment this line to view only cited bib entries;

\bibliography{Bibliography-APA}%

\begin{thebibliography}{}

\bibitem[Adeli, 2001]{adeli2001neural}
Adeli, H. (2001).
\newblock Neural networks in civil engineering: 1989--2000.
\newblock {\em Computer-Aided Civil and Infrastructure Engineering},
  16(2):126--142.

\bibitem[Alfarrarjeh et~al., 2018]{alfarrarjeh2018deep}
Alfarrarjeh, A., Trivedi, D., Kim, S.~H., and Shahabi, C. (2018).
\newblock A deep learning approach for road damage detection from smartphone
  images.
\newblock In {\em 2018 IEEE International Conference on Big Data (Big Data)},
  pages 5201--5204. IEEE.

\bibitem[Anand et~al., 2018]{anand2018crack}
Anand, S., Gupta, S., Darbari, V., and Kohli, S. (2018).
\newblock Crack-pot: Autonomous road crack and pothole detection.
\newblock In {\em 2018 Digital Image Computing: Techniques and Applications
  (DICTA)}, pages 1--6. IEEE.

\bibitem[Angulo et~al., 2019]{angulo2019road}
Angulo, A., Vega-Fern{\'a}ndez, J.~A., Aguilar-Lobo, L.~M., Natraj, S., and
  Ochoa-Ruiz, G. (2019).
\newblock Road damage detection acquisition system based on deep neural
  networks for physical asset management.
\newblock In {\em Mexican International Conference on Artificial Intelligence},
  pages 3--14. Springer.

\bibitem[Budzar, 2018]{Budzar2018}
Budzar, M. (2018).
\newblock Re-training the model with images using tensorflow.
\newblock Retrieved from:
  \url{https://proandroiddev.com/re-training-the-model-with-images-using-tensorflow-7758e9eb8db5}.

\bibitem[Casas-Avellaneda and L{\'o}pez-Parra, 2016]{casas2016detection}
Casas-Avellaneda, D.~A. and L{\'o}pez-Parra, J.~F. (2016).
\newblock Detection and localization of potholes in roadways using smartphones.
\newblock {\em Dyna}, 83(195):156--162.

\bibitem[Cha et~al., 2018]{cha2018autonomous}
Cha, Y.-J., Choi, W., Suh, G., Mahmoudkhani, S., and
  B{\"u}y{\"u}k{\"o}zt{\"u}rk, O. (2018).
\newblock Autonomous structural visual inspection using region-based deep
  learning for detecting multiple damage types.
\newblock {\em Computer-Aided Civil and Infrastructure Engineering},
  33(9):731--747.

\bibitem[Dertat, 2017]{Dertat2017}
Dertat, A. (2017).
\newblock Applied deep learning – part 4: Convolutional neural networks.
\newblock Retrieved from:
  \url{https://towardsdatascience.com/applied-deep-learning-part-4-convolutional-neural-networks-584bc134c1e2}.

\bibitem[Douillard, 2018]{Douillard2018}
Douillard, A. (2018).
\newblock 3 small but powerful convolutional networks.
\newblock Retrieved from: \url{https://towardsdatascience.com
  /3-small-but-powerful-convolutional-networks-27ef86faa42d}.

\bibitem[Du et~al., 2020]{du2020pavement}
Du, Y., Pan, N., Xu, Z., Deng, F., Shen, Y., and Kang, H. (2020).
\newblock Pavement distress detection and classification based on yolo network.
\newblock {\em International Journal of Pavement Engineering}, pages 1--14.

\bibitem[Everingham et~al., 2010]{everingham2010pascal}
Everingham, M., Van~Gool, L., Williams, C.~K., Winn, J., and Zisserman, A.
  (2010).
\newblock The pascal visual object classes (voc) challenge.
\newblock {\em International journal of computer vision}, 88(2):303--338.

\bibitem[Fan et~al., 2018]{fan2018automatic}
Fan, Z., Wu, Y., Lu, J., and Li, W. (2018).
\newblock Automatic pavement crack detection based on structured prediction
  with the convolutional neural network.
\newblock {\em arXiv preprint arXiv:1802.02208}.

\bibitem[Girshick, 2015]{girshick2015fast}
Girshick, R. (2015).
\newblock Fast r-cnn.
\newblock In {\em Proceedings of the IEEE international conference on computer
  vision}, pages 1440--1448.

\bibitem[Girshick et~al., 2014]{girshick2014rich}
Girshick, R., Donahue, J., Darrell, T., and Malik, J. (2014).
\newblock Rich feature hierarchies for accurate object detection and semantic
  segmentation.
\newblock In {\em Proceedings of the IEEE conference on computer vision and
  pattern recognition}, pages 580--587.

\bibitem[Goodfellow et~al., 2016]{goodfellow2016deep}
Goodfellow, I., Bengio, Y., and Courville, A. (2016).
\newblock {\em Deep learning}.
\newblock MIT press.

\bibitem[Hasenauer et~al., 2001]{hasenauer2001estimating}
Hasenauer, H., Merkl, D., and Weingartner, M. (2001).
\newblock Estimating tree mortality of norway spruce stands with neural
  networks.
\newblock {\em Advances in Environmental Research}, 5(4):405--414.

\bibitem[He et~al., 2016]{he2016deep}
He, K., Zhang, X., Ren, S., and Sun, J. (2016).
\newblock Deep residual learning for image recognition.
\newblock In {\em Proceedings of the IEEE conference on computer vision and
  pattern recognition}, pages 770--778.

\bibitem[Howard et~al., 2017]{howard2017mobilenets}
Howard, A.~G., Zhu, M., Chen, B., Kalenichenko, D., Wang, W., Weyand, T.,
  Andreetto, M., and Adam, H. (2017).
\newblock Mobilenets: Efficient convolutional neural networks for mobile vision
  applications.
\newblock {\em arXiv preprint arXiv:1704.04861}.

\bibitem[Huang et~al., 2017]{huang2017speed}
Huang, J., Rathod, V., Sun, C., Zhu, M., Korattikara, A., Fathi, A., Fischer,
  I., Wojna, Z., Song, Y., Guadarrama, S., et~al. (2017).
\newblock Speed/accuracy trade-offs for modern convolutional object detectors.
\newblock In {\em Proceedings of the IEEE conference on computer vision and
  pattern recognition}, pages 7310--7311.

\bibitem[Intel, 2018]{Intel2018}
Intel (2018).
\newblock Pedestrian detection using tensorflow on intel architecture (white
  paper), intel ai builders.
\newblock Retrieved from:
  \url{https://builders.intel.com/docs/aibuilders/pedestrian-detection-using-tensorflow-on-intel-architecture.pdf}.

\bibitem[IRC:82, 2015]{IRC82_2015}
IRC:82 (2015).
\newblock Code of practice for maintenance of bituminous road surfaces.
\newblock Technical report, Indian Road Congress, New Delhi, India.

\bibitem[Kargah-Ostadi et~al., 2017]{kargah2017evaluation}
Kargah-Ostadi, N., Nazef, A., Daleiden, J., and Zhou, Y. (2017).
\newblock Evaluation framework for automated pavement distress identification
  and quantification applications.
\newblock {\em Transportation Research Record}, 2639(1):46--54.

\bibitem[Kluger et~al., 2018]{kluger2018region}
Kluger, F., Reinders, C., Raetz, K., Schelske, P., Wandt, B., Ackermann, H.,
  and Rosenhahn, B. (2018).
\newblock Region-based cycle-consistent data augmentation for object detection.
\newblock In {\em 2018 IEEE International Conference on Big Data (Big Data)},
  pages 5205--5211. IEEE.

\bibitem[Lin et~al., 2017]{lin2017structural}
Lin, Y.-z., Nie, Z.-h., and Ma, H.-w. (2017).
\newblock Structural damage detection with automatic feature-extraction through
  deep learning.
\newblock {\em Computer-Aided Civil and Infrastructure Engineering},
  32(12):1025--1046.

\bibitem[Liu et~al., 2016]{liu2016ssd}
Liu, W., Anguelov, D., Erhan, D., Szegedy, C., Reed, S., Fu, C.-Y., and Berg,
  A.~C. (2016).
\newblock Ssd: Single shot multibox detector.
\newblock In {\em European conference on computer vision}, pages 21--37.
  Springer.

\bibitem[Maeda et~al., 2020]{maeda2020}
Maeda, H., Kashiyama, T., Sekimoto, Y., Seto, T., and Omata, H. (2020).
\newblock Generative adversarial network for road damage detection.
\newblock {\em Computer-Aided Civil and Infrastructure Engineering}.

\bibitem[Maeda et~al., 2018]{maeda2018road}
Maeda, H., Sekimoto, Y., Seto, T., Kashiyama, T., and Omata, H. (2018).
\newblock Road damage detection and classification using deep neural networks
  with smartphone images.
\newblock {\em Computer-Aided Civil and Infrastructure Engineering},
  33(12):1127--1141.

\bibitem[Majidifard et~al., 2020]{majidifard2020pavement}
Majidifard, H., Jin, P., Adu-Gyamfi, Y., and Buttlar, W.~G. (2020).
\newblock Pavement image datasets: A new benchmark dataset to classify and
  densify pavement distresses.
\newblock {\em Transportation Research Record}, 2674(2):328--339.

\bibitem[McGhee, 2004]{mcghee2004automated}
McGhee, K. (2004).
\newblock Automated pavement distress collection techniques. nchrp synthesis
  334. national cooperative highway research program.
\newblock {\em Transportation Research Board, Washington DC}.

\bibitem[Mertz et~al., 2014]{mertz2014city}
Mertz, C., Varadharajan, S., Jose, S., Sharma, K., Wander, L., and Wang, J.
  (2014).
\newblock City-wide road distress monitoring with smartphones.
\newblock In {\em Proceedings of ITS World Congress}, pages 1--9.

\bibitem[Miller and Zaloshnja, 2009]{miller2009crash}
Miller, T.~R. and Zaloshnja, E. (2009).
\newblock On a crash course: The dangers and health costs of deficient
  roadways.

\bibitem[Peppa et~al., 2018]{peppa2018urban}
Peppa, M., Bell, D., Komar, T., and Xiao, W. (2018).
\newblock Urban traffic flow analysis based on deep learning car detection from
  cctv image series.
\newblock In {\em SPRS TC IV Mid-term Symposium “3D Spatial Information
  Science--The Engine of Change”}. Newcastle University.

\bibitem[Pierce et~al., 2013]{pierce2013practical}
Pierce, L.~M., McGovern, G., and Zimmerman, K.~A. (2013).
\newblock Practical guide for quality management of pavement condition data
  collection.

\bibitem[Prabhu, 2018]{prabhu2018understanding}
Prabhu, R. (2018).
\newblock Understanding of convolutional neural network (cnn)--deep learning.
\newblock {\em A Medium Corporation, US}.

\bibitem[Radopoulou and Brilakis, 2015]{radopoulou2015detection}
Radopoulou, S.~C. and Brilakis, I. (2015).
\newblock Detection of multiple road defects for pavement condition assessment.
\newblock {\em Eindhoven, The Netherlands}.

\bibitem[Ren et~al., 2015]{ren2015faster}
Ren, S., He, K., Girshick, R., and Sun, J. (2015).
\newblock Faster r-cnn: Towards real-time object detection with region proposal
  networks.
\newblock In {\em Advances in neural information processing systems}, pages
  91--99.

\bibitem[Roadware, 2019]{Fugro2019}
Roadware, F. (2019).
\newblock Pavement condition assessment, data sheets.
\newblock Retrieved from:
  \url{http://www.fugroroadware.com/related/english-alldatasheets}.

\bibitem[Roberts et~al., 2020]{roberts2020towards}
Roberts, R., Giancontieri, G., Inzerillo, L., and Di~Mino, G. (2020).
\newblock Towards low-cost pavement condition health monitoring and analysis
  using deep learning.
\newblock {\em Applied Sciences}, 10(1):319.

\bibitem[Sharma, 2017]{Sharma2017}
Sharma, S. (2017).
\newblock Activation functions: Neural networks.
\newblock Retrieved from: \url{https://towardsdatascience.com/
  activation-functions-neural-networks-1cbd9f8d91d6}.

\bibitem[Silva and Lucena, 2018]{LealDeSilva2018concrete}
Silva, W. R. L.~d. and Lucena, D. S.~d. (2018).
\newblock Concrete cracks detection based on deep learning image
  classification.
\newblock In {\em Multidisciplinary Digital Publishing Institute Proceedings},
  volume~2, page 489.

\bibitem[Szegedy et~al., 2016]{szegedy2016rethinking}
Szegedy, C., Vanhoucke, V., Ioffe, S., Shlens, J., and Wojna, Z. (2016).
\newblock Rethinking the inception architecture for computer vision.
\newblock In {\em Proceedings of the IEEE conference on computer vision and
  pattern recognition}, pages 2818--2826.

\bibitem[Wang et~al., 2018a]{wang2018road}
Wang, W., Wu, B., Yang, S., and Wang, Z. (2018a).
\newblock Road damage detection and classification with faster r-cnn.
\newblock In {\em 2018 IEEE International Conference on Big Data (Big Data)},
  pages 5220--5223. IEEE.

\bibitem[Wang et~al., 2018b]{wang2018deep}
Wang, Y.~J., Ding, M., Kan, S., Zhang, S., and Lu, C. (2018b).
\newblock Deep proposal and detection networks for road damage detection and
  classification.
\newblock In {\em 2018 IEEE International Conference on Big Data (Big Data)},
  pages 5224--5227. IEEE.

\bibitem[Zalama et~al., 2014]{zalama2014road}
Zalama, E., G{\'o}mez-Garc{\'\i}a-Bermejo, J., Medina, R., and Llamas, J.
  (2014).
\newblock Road crack detection using visual features extracted by gabor
  filters.
\newblock {\em Computer-Aided Civil and Infrastructure Engineering},
  29(5):342--358.

\bibitem[Zhang et~al., 2017]{zhang2017automated}
Zhang, A., Wang, K.~C., Li, B., Yang, E., Dai, X., Peng, Y., Fei, Y., Liu, Y.,
  Li, J.~Q., and Chen, C. (2017).
\newblock Automated pixel-level pavement crack detection on 3d asphalt surfaces
  using a deep-learning network.
\newblock {\em Computer-Aided Civil and Infrastructure Engineering},
  32(10):805--819.

\bibitem[Zhang et~al., 2016]{zhang2016road}
Zhang, L., Yang, F., Zhang, Y.~D., and Zhu, Y.~J. (2016).
\newblock Road crack detection using deep convolutional neural network.
\newblock In {\em Image Processing (ICIP), 2016 IEEE International Conference
  on}, pages 3708--3712. IEEE.

\end{thebibliography}
\bibliographystyle{apalike}

\end{document}